\documentclass[sigconf]{acmart}
\usepackage{balance}
\usepackage{soul}
\usepackage{url}
\usepackage{graphicx}
\usepackage{amsmath}
\usepackage{booktabs}
\usepackage{amsthm}
\usepackage{subfig}
\usepackage{algorithm}
\usepackage{algorithmic}
\usepackage{multirow}

\newcommand{\defeq}{\mathrel{\mathop:}=}
\AtBeginDocument{%
  }

\copyrightyear{2023}
\acmYear{2023}
\setcopyright{acmlicensed}\acmConference[MM '23]{Proceedings of the 31st ACM International Conference on Multimedia}{October 29-November 3, 2023}{Ottawa, ON, Canada}
\acmBooktitle{Proceedings of the 31st ACM International Conference on Multimedia (MM '23), October 29-November 3, 2023, Ottawa, ON, Canada}
\acmPrice{15.00}
\acmDOI{10.1145/3581783.3612099}
\acmISBN{979-8-4007-0108-5/23/10}

\begin{document}

\title{Practical Edge Detection via Robust Collaborative Learning}

\author{Yuanbin Fu}
\affiliation{%
  \institution{Tianjin University}
  \country{Tianjin, China}
}
\email{yuanbinfu@tju.edu.cn}

\author{Xiaojie Guo}
\authornote{Corresponding author.}
\affiliation{%
  \institution{Tianjin University}
  \country{Tianjin, China}
}
\email{xj.max.guo@gmail.com}


\begin{abstract}
Edge detection, as a core component in a wide range of vision-oriented tasks, is to identify object boundaries and prominent edges in natural images. An edge detector is desired to be both efficient and accurate for practical use. To achieve the goal, two key issues should be concerned: 1) How to liberate deep edge models from inefficient pre-trained backbones that are leveraged by most existing deep learning methods, for saving the computational cost and cutting the model size; and 2) How to mitigate the negative influence from noisy or even wrong labels in training data, which widely exist in edge detection due to the subjectivity and ambiguity of annotators, for the robustness and accuracy. In this paper, we attempt to simultaneously address the above problems via developing a collaborative learning based model, termed PEdger. The principle behind our PEdger is that, the information learned from different training moments and heterogeneous (recurrent and non recurrent in this work) architectures, can be assembled to explore robust knowledge against noisy annotations, even without the help of pre-training on extra data. Extensive ablation studies together with quantitative and qualitative experimental comparisons on the BSDS500 and NYUD datasets are conducted to verify the effectiveness of our design, and demonstrate its superiority over other competitors in terms of accuracy, speed, and model size. Codes can be found at {\url{https://github.com/ForawardStar/PEdger}}.
\end{abstract}

\begin{teaserfigure}
\centering
   \subfloat[Input \& GT]{\includegraphics[width=0.195\linewidth]{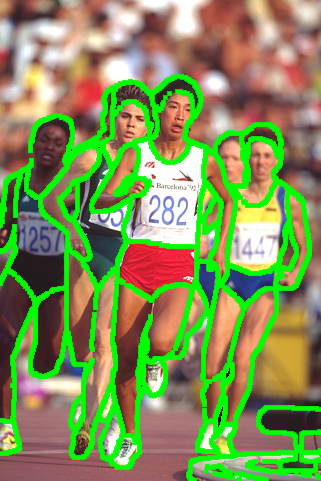}}
   \subfloat[RCF]{\includegraphics[width=0.195\linewidth]{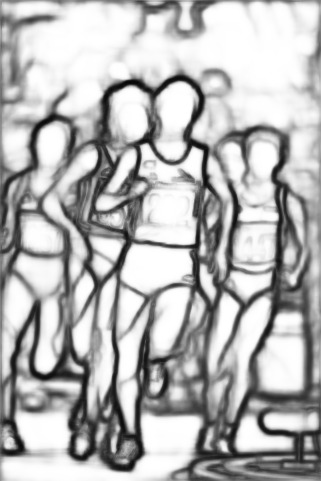}}
   \subfloat[PiDiNet]{\includegraphics[width=0.195\linewidth]{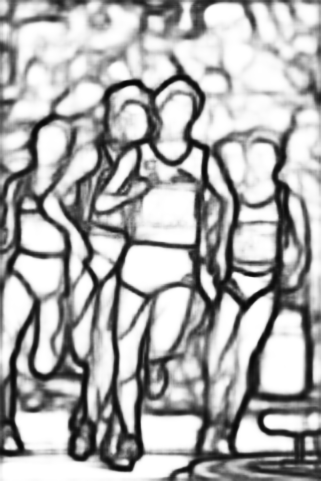}}
   \subfloat[BDCN]{\includegraphics[width=0.195\linewidth]{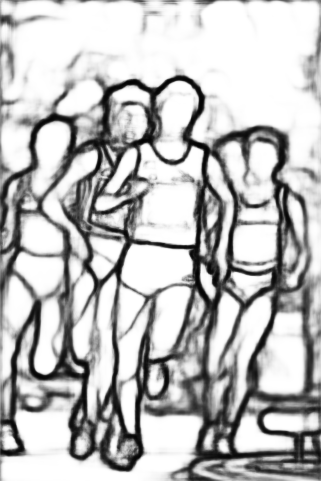}}
   \subfloat[Ours]{\includegraphics[width=0.195\linewidth]{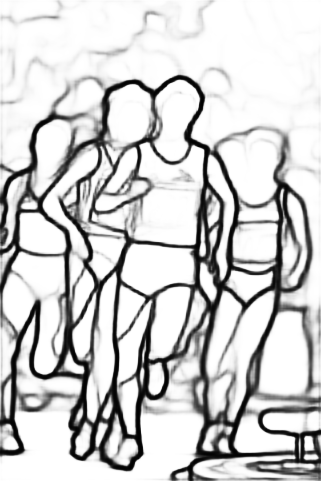}}
  \caption{Visual comparison with other methods. Our detected edges are thinner and clearer than the others.}
  \label{example}
\end{teaserfigure}
\maketitle

\section{Introduction}
\label{sec:intro}
 Edge detection aims to extract object boundaries and salient edges from given images, please see Fig.~\ref{example} for an example. As an indispensable component to a wide spectrum of applications, such as object detection \cite{faster-rcnn}, semantic segmentation \cite{JSENet}, and image editing \cite{editing}, to just name a few, this task has been drawing intensive attention from the community with a great progress made over last decades, especially with the emergence of deep learning. Even though, by simultaneously considering the detection accuracy, processing speed, and model size, developing practical edge detectors for real-time applications remains highly desired. 
 
\begin{figure}[t]
    \centering
    \includegraphics[width=\linewidth]{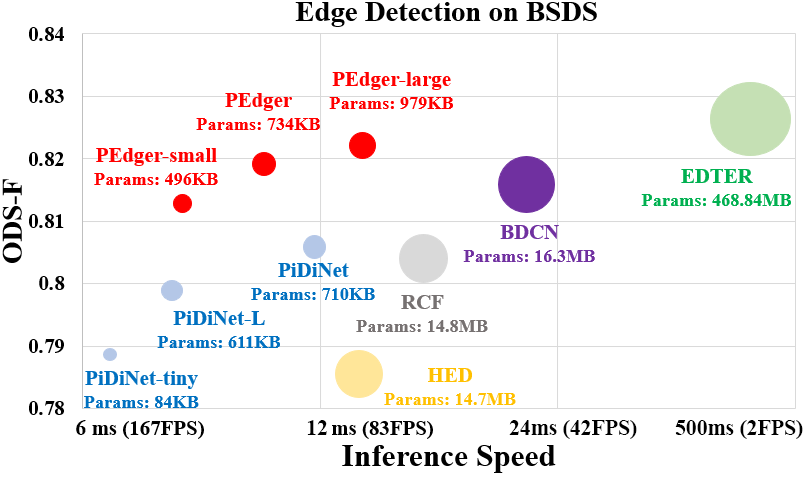}
    \caption{Performance comparison in Optimal Dataset Scale F-score (ODS-F). For all the methods, the running speed is tested on an RTX 3090 GPU with $200\times 200$ images as input.}
    \label{model-comp}
\end{figure}

\begin{figure}[t]
    \centering

    \includegraphics[width=0.24\linewidth]{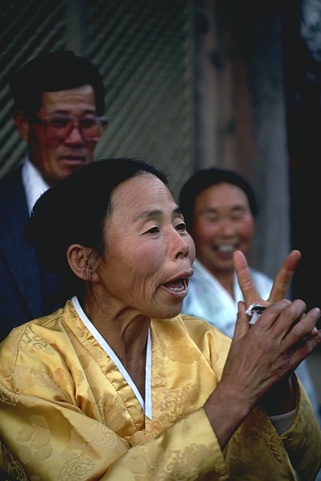}
    \includegraphics[width=0.24\linewidth]{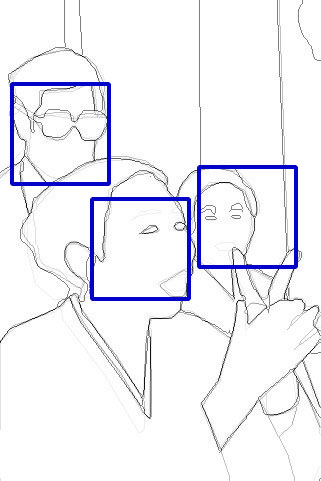}
    \includegraphics[width=0.24\linewidth]{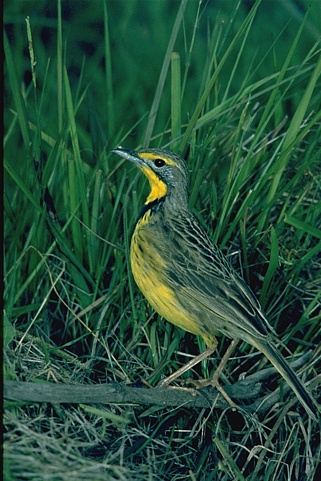}
    \includegraphics[width=0.24\linewidth]{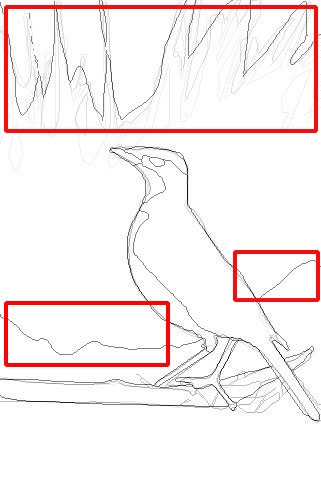}

     \quad Image1 \quad\quad Annotation1 \quad\quad Image2 \quad\quad Annotation2 \quad
    
    \caption{Some examples of noisy annotations in BSDS. The labeling of eyes and mouth indicated by the blue squares in the first case is contradictory. The red squares in the second case show the wrong annotations that some pixels within the grass are labelled as edges, which actually are textures.}
    \label{noisysamples}
\end{figure}

Early edge detectors concentrated on local features, like image gradients \cite{Canny} and manually-designed features \cite{pb,gpu-ucm,se}. These methods determine edges solely based on shallow/low-level cues, the accuracy of which is often unsatisfactory in practice. To exploit semantic information, numerous  works \cite{HED,BDCN_cvpr,RCF,EDTER} have been recently built on top of deep learning techniques and have brought tremendous progresses, benefiting from the strong representative ability of deep networks. These approaches leveraged the models like VGG16 \cite{VGG16}, ResNet50 \cite{resnet}, and ViT-B/16 \cite{vit} pre-trained on the large-scale ImageNet \cite{ImageNet}, as their backbones, to embody abundant knowledge. In spite of the remarkable improvement in terms of accuracy, \textit{they inevitably pay high prices at computation and memory, due to the (large) pre-trained backbones}. The inefficiency limits their applicability in real-time scenarios. In the literature, a good trade-off technique goes to the recent PiDiNet \cite{PiDiNet}, which liberates from heavy pre-trained backbones, and achieves fast and reasonably accurate performance. Even though, there still has a large room for improving the accuracy, as illustrated in Fig.~\ref{model-comp}. An important issue rarely considered by these mentioned methods is \textit{the negative impact from noisy or even wrong annotations in training data on the quality of learning. Particularly in edge detection datasets, such problematic labels widely exist due to the subjectivity and ambiguity of annotators.} Figure \ref{noisysamples} depicts  noisy annotations.

To take a step forward in deep learning era for accurate, fast, and lightweight edge detection, as previously analyzed, we have to face two main challenges.
On the one hand, the high accuracy of most deep models is primarily attributed to the abundant knowledge implicated in (large) pre-trained parameters, but these pre-trained parameters significantly increase the computational complexity and model size. A question naturally arises, \textit{i.e., how to liberate deep edge models from pre-trained backbones for saving the computational cost and cutting the model size?} On the other hand, for a given image, different annotators may provide nonidentical edge labels. In other words, the inconsistent/noisy labels by ambiguous and subjective annotators can hardly be avoided. Hence, another challenge, \textit{i.e., how to mitigate the negative influence from noisy or even wrong labels in training data for the robustness and accuracy}, needs to be solved. It seems that pre-trained models can reduce the influence of noisy labels  thanks to the regularization ability from extra data, but it again falls into the first dilemma. 

For jointly conquering the above two challenges, we alternatively investigate to integrate useful knowledge, probably weak, from different sources (except for large pre-trained models) as a strong agent. By this means, the bias/inconsistency of the training data can be relieved from the perspective of ensemble learning \cite{Heterogeneous,SELC}, while dispensing with the pre-training. Although the knowledge learned either at a particular training moment or by a single model is easily disturbed by noisy/outlier samples, especially when the batch size is small, it shall be qualified to act as a source for ensemble. To verify the primary claim of this work, the information from both different training moments and heterogeneous network architectures, is taken into consideration.

\paragraph{Contribution.}
With regard to heterogeneous architectures, two networks with recurrent (slower) and non recurrent (faster) structures are incorporated during training. The knowledge learned by these two structures will be fused using a simple uncertainty-aware weighting method.  For integrating the information across different training moments, we ensemble the network parameters corresponding to different epochs in a momentum way. During training, both the networks are updated collaboratively, that is, they are trained in one stage without waiting for the training of another to be finished beforehand. Please notice that in the testing phase, only one network needs to be executed.
Our major contributions are summarized as:
\begin{itemize}
    \item  We customize a collaborative learning scheme for practical edge detection via assembling the information across different network architectures and training moments. It is able to prevent the network from over-fitting to noises without the help of any pre-trained model.
    \item We design two architectures to capture diversified knowledge for robustness, which will be fused in an uncertainty-aware fashion. Particularly, a recurrent structure with shared parameters across scales and a non-recurrent one having different amount of parameters for different scales, are maintained in the collaborative learning process.
    \item  We conduct extensive experiments to verify the effectiveness of our design on several benchmark datasets. The results indicate that our method can achieve superior performance over other state-of-the-art competitors by comprehensively considering the accuracy, speed, and model size.
\end{itemize}  

\section{Related Work}
\label{sec:related}
\paragraph{Edge Detection.} 
Early attempts on edge detection may trace back to gradient based edge operators like Canny \cite{Canny}. Though being efficient, the detection quality by these operators hardly reaches the requirement in real-world applications. Another category refers to machine learning approaches. As a representative, gpb-UCM \cite{gpu-ucm} feeds the oriented gradient of three brightness channels and one texture channel as the features into a logistic regression classifier. Dollar \textit{et al.} \cite{se} proposed to employ the random forest for edge detection, which takes full advantage of the inherent structure in local patches. 
With the emergence of deep learning, various approaches have been proposed, such as HED \cite{HED} and RCF \cite{RCF}. They output multi-scale side-outputs based on the pre-trained features from VGG16. But, all the side-outputs are trained under identical supervision, ignoring the diversity. To address this issue, several works \cite{relaxed,BDCN_cvpr} advocated to impose different supervisions on these side-outputs. More recently, EDTER \cite{EDTER} accomplishes the edge detection task using ViT \cite{vit}. Although considerably improving the accuracy, these mentioned deep learning based methods all employ external networks (\textit{e.g.}, VGG16, ViT-B/16) with substantial parameters pre-trained on large-scale datasets (\textit{e.g.} ImageNet \cite{ImageNet}), as assistant. Consequently, their computational costs also increase. To save the computational expense, PiDiNet \cite{PiDiNet} introduces the pixel difference convolution to extract edge-related features without employing pre-trained backbones. Compared to RCF \cite{RCF}, it considerably reduces the model size and accelerates the speed with competitive accuracy.

\paragraph{Learning from Noisy Labels.}
Recently, numerous approaches have been raised to train deep networks in the presence of label noises \cite{teachingwithsoft,PresonRID,symmetricEntropy,ParkJBHJKP20}. Among them, rectifying the original noisy targets based on the useful information/knowledge investigated in training, is demonstrated to be effective. For example,  PENCIL \cite{DBLP:conf/cvpr/YiW19} adopts the information in back-propagation to probabilistically correct noisy labels, which does not need any prior information about noises. Similarly, Ren \textit{et al.} \cite{DBLP:conf/icml/RenZYU18} estimated the importance score of samples according to gradient directions in back-propagation.

Besides the information from back-propagation, softening the noisy targets using the knowledge learned from previous iterations, \textit{i.e.}, online knowledge distillation, is also revered by many works. As representatives, Kim \textit{et al.}  \cite{SelfDistillation} produced the soft target based on the predictions of the network itself, using the parameters updated at the last iteration. \cite{mutual} utilized peer-teaching strategy that trains collaboratively multiple networks to learn from each other. But, one independent network may not provide reasonable guidance. Hence, \cite{ONE,OnlineDis} established the supervision information on-the-fly through ensembling the predictions of multiple collaborative networks, which achieve better performance. The methods in this category suffer from two main limitations: 1) The network architectures  are identical or similar, leading to the poor diversity; 2) They utilize the parameters updated by the last mini-batch to generate soft targets, resulting in the inconsistent optimization directions since the parameters change dramatically for each iteration.

All the above mentioned researches point at image-level noisy annotations for classification, which may not be suitable for pixel-level noises. For dealing with dense prediction tasks, \cite{DBLP:conf/cvpr/FengLZYWM22,DBLP:journals/pami/ZhangDZHBH21} assigned a pixel-wise soft weight to each sample, which is determined by the variance of predictions at different iterations, for the sake of encouraging noisy pixels to contribute less than other pixels.  Unfortunately, besides involving only one homogeneous network structure during training, their predictions are produced only based on the snapshot knowledge at each iteration, without the consideration of knowledge across different training moments. To mitigate the pixel-level noises in edge detection datasets, we take advantage of the robust knowledge from different training moments and heterogeneous models, where the poor diversity of models and inconsistent label constraints during training, can be avoided.

\section{Methodology}
This work aims to devise a practical edge detector by simultaneously addressing 1) the dependence on large pre-trained models, and 2) the negative influence of noisy labels in training data. To achieve the goal, we propose to integrate the knowledge across different training moments and network architectures (as schematically shown in Fig. \ref{arch}).  We denote the training data by $\textbf{D} \defeq \{(X_n, Y_n), n = 1,2, ..., N\}$, where $N$ is the total number of training samples, each sample $X_n$ is the raw input image, and $Y_n$ is the ground truth edge map for $X_n$.

\subsection{Network Architecture}
\label{NetworkArchitecture}
Our framework comprises recurrent and non recurrent networks with different randomly initialized parameters, to model diversified features for robustness, which are trained collaboratively in one stage. It is rational since integrating the predictions from discrepant models, helps to prevent from over-fitting to noises/outliers, as verified by the literature of model ensemble \cite{Heterogeneous,SELC}. During inference, only the non-recurrent structure is executed. 

\begin{figure}[t]
    \centering
    \subfloat[Recurrent Architecture.]{
    \includegraphics[width=0.9\linewidth]{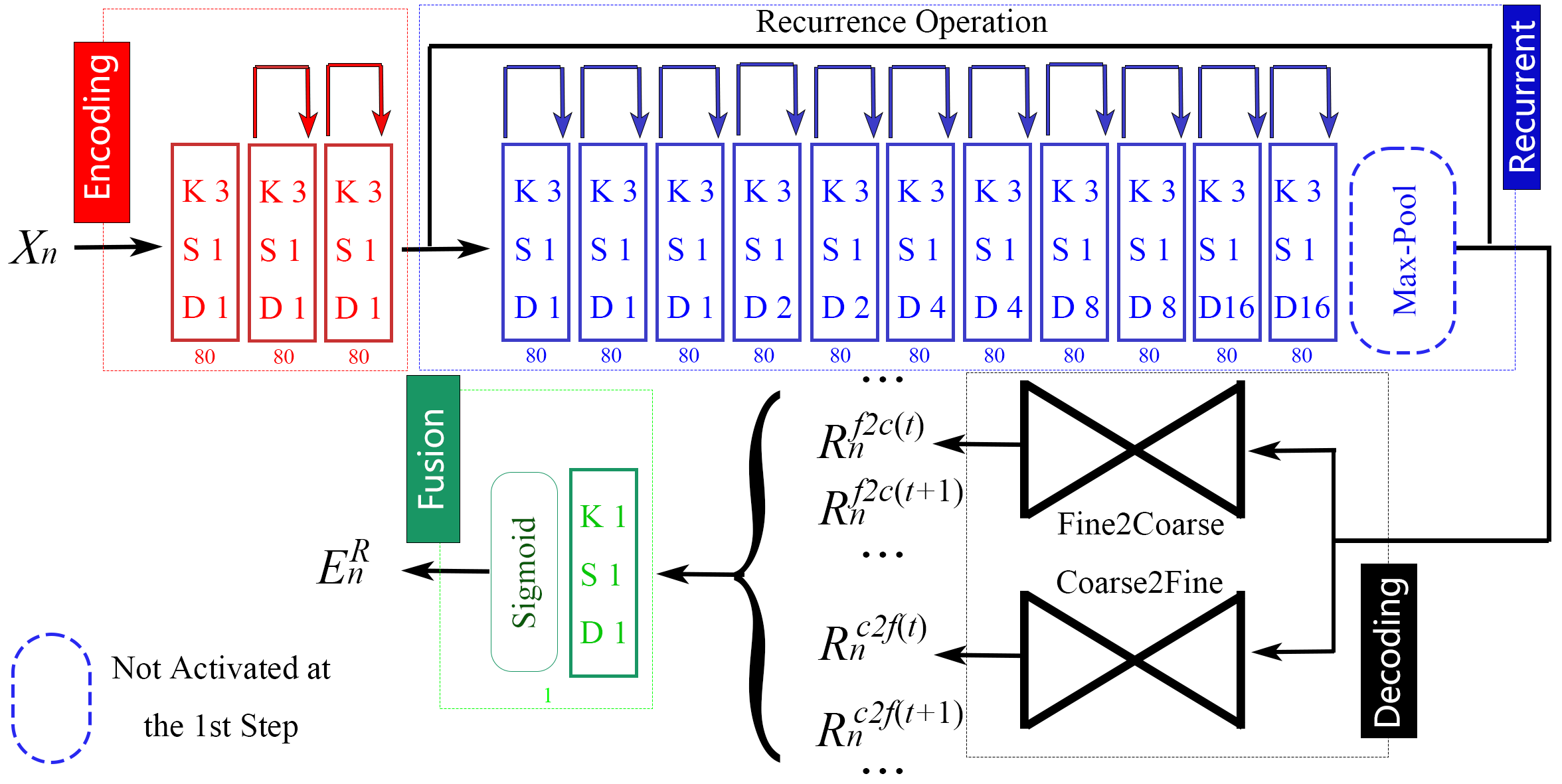}
    }
    
    \subfloat[Non Recurrent Architecture.]{
    \includegraphics[width=0.9\linewidth]{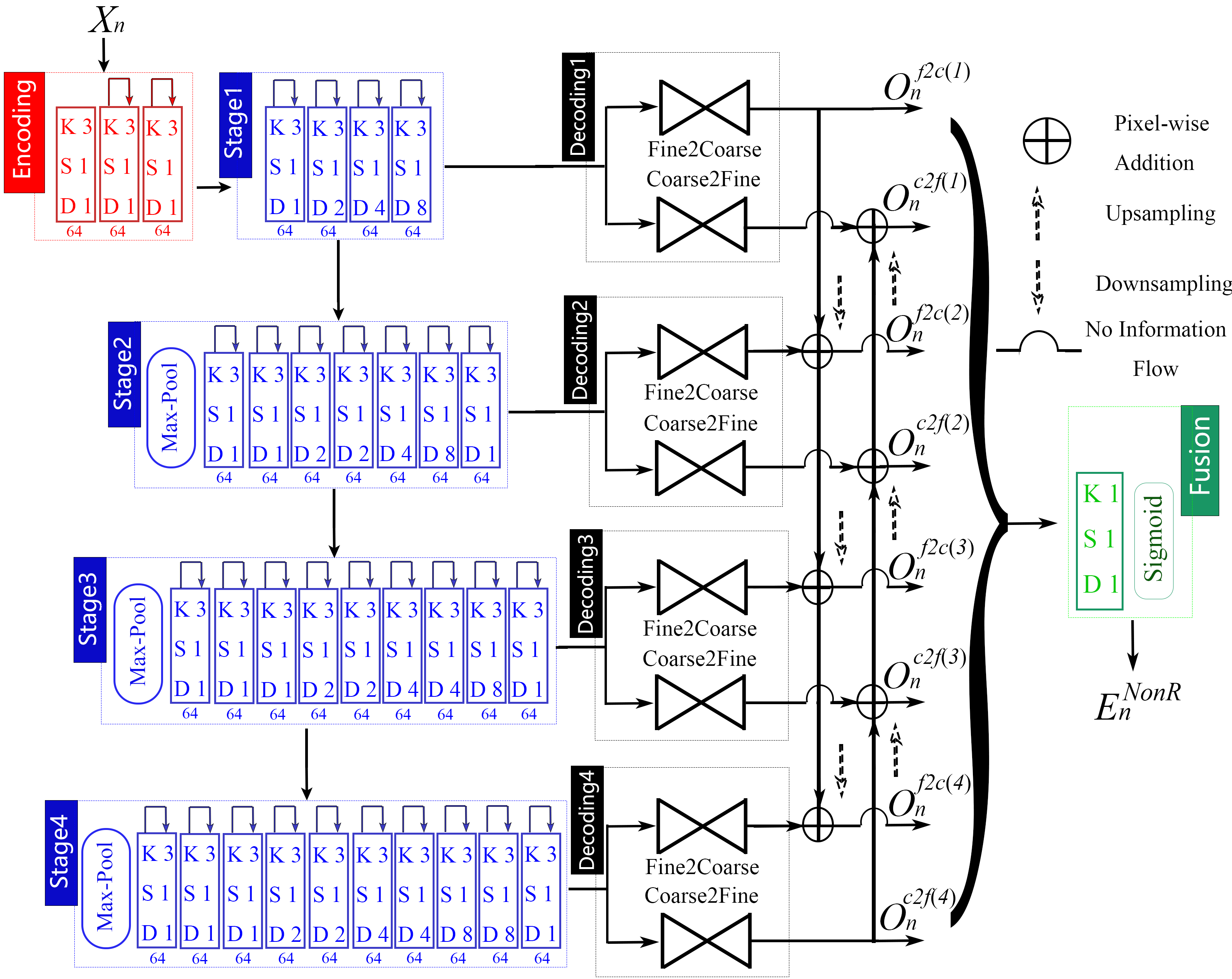}
    }
    
    \caption{Recurrent \& non recurrent architecture. $K$, $S$, and $D$ are the kernel size, stride, and dilation rate, respectively. The number of output channels is shown below each block.}
    \label{arch}
\end{figure}

\paragraph{Recurrent Architecture.} The blueprint of our recurrent network is shown in Fig. \ref{arch} (a). For each data sample $X_n$, it is firstly taken as the input of the encoding module consisting of one plain convolutional layer and two residual layers. The features extracted by the encoding module will be fed into the recurrent module whose parameters are shared across every recurrent steps. Specifically, at $t$-th step ($t > 1$), the recurrent module takes the features produced by the $(t-1)$-th step as input (for the first recurrent step, that is $t = 1$, the features generated by the encoding module are inputted), and then outputs the features of this step that will be taken as the input of next step again. This process is executed iteratively until reaching the total recurrent step $T$. 

Notice that the recurrent module is connected with a max-pooling operation at every steps except for the first step. As demonstrated by \cite{HED,RCF,BDCN_cvpr}, the usage of max-pooling operation can increase the receptive fields, which is beneficial to generate multi-scale side-output edges. The useful information captured by each step becomes coarser as its receptive field size increasing, and the responses at larger objects are stronger. 
In addition, inspired by the works \cite{LPCB,CED,BDCN_cvpr} that aggregate the information from fine-to-coarse and coarse-to-fine for edge detection, we introduce two branches in the decoding module as shown in Fig. \ref{arch}, to aggregate the information from the early recurrent steps (fine) to late steps (coarse), and from the late steps to early, respectively. Concretely, for each step $t$, the side-output features $R_n^{f2c(t)}$ at $t$-th step are obtained through adding the outputs by the fine-to-coarse branch of this step, to the down-sampled side-output features of previous step, say $R_n^{f2c(t-1)}$. Similarly, $R_n^{c2f(t)}$ is obtained through adding the outputs by the coarse-to-fine branch of this step, to the up-sampled features $R_n^{c2f(t+1)}$ of the late step. Finally, all the side-output features are fused by a $1\times 1$ convolution, and processed by the Sigmoid activation to output the edge detection results $E_n^R$. Due to the shared parameters, the recurrent network is compact and converges fast. Also, the number of channels in recurrent model is larger than the non recurrent one, thanks to that the shared parameters can save parameter space for expanding channels.

\paragraph{Non-recurrent Architecture.} Objectively, recurrent models are typically slow in speed, because the recurrent module (shown in blue part of Fig.\ref{arch} (a)) needs to be inferred for many times, where the network configurations are fixed.  To construct a faster network, we propose to train different amount of parameters specific to different scales (shown in blue part of Fig.\ref{arch} (b)). Specifically, the parameter amount of the module at the first stage is the least, and gradually increases as the feature size decreases. The reason for this design is that, the earlier stages mainly focus on extracting lower-level features with larger resolutions, while the later stages produce higher-level semantic features with smaller resolutions. In short, the higher-level semantic structures at later stages require more parameters to learn than lower-level details, and it takes more time to process the feature maps at later stages with larger resolutions than feature maps with smaller resolutions. Thus, using lesser (more) parameters for earlier (later) stages can achieve minimal performance degradation and maximal speed acceleration.
It can be seen from Fig. \ref{arch} that the difference between the recurrent and non recurrent structure is whether the parameters of the modules indicated by the blue and black (the decoding module) are shared across multiple scales. Notice that the non recurrent architecture also adopts the bidirectional aggregation, the operation of which is exactly the same as the recurrent one.

\begin{algorithm}[tb]
    \caption{Collaborative Learning for Edge Detection}
    \label{alg:algorithm}
    \raggedright
    \textbf{Input}: Training dataset $\textbf{D}$; the total number of samples $N$; the total number of training epochs $EP$; two network  $G^R$ and $G^{NonR}$ trained by back-propagation; two momentum networks $G_m^R$ and $G_m^{NonR}$; a hyper-parameter $\eta_{EP}$ \\
    \textbf{Output}: Updated  target network $G_m^{NonR}$\\
    \textbf{Initialize}: $ep\defeq0$; $\eta_{EP}\defeq0.8$; Randomly initialize $G^R$ and $G^{NonR}$
    \begin{algorithmic}[1] 
        \WHILE{$ep < EP$}
        \STATE Compute $\eta_{ep}$ by Eq. \eqref{weightincrease};
         \WHILE{$n < N$}
        \STATE Sample data $X_n$, $Y_n$ from $\textbf{D}$;
        \IF {$ep > 0$}
        \STATE $M_n^R\defeq G_m^R(X_n)$, $M_n^{NonR}\defeq G_m^{NonR}(X_n)$;
        \STATE Compute $M_n$ by Eq. \eqref{fusion} and $\widetilde{Y}_n$ by Eq. \eqref{soft};
        \ELSE
        \STATE  $\widetilde{Y}_n \defeq Y_n$
        \ENDIF
        \STATE $E_n^R\defeq G^R(X_n)$, $E_n^{NonR}\defeq G^{NonR}(X_n)$;
        \STATE Compute $L^R$ by Eq. \eqref{loss1} and $L^{NonR}$ by Eq. \eqref{loss2};
        \STATE Update $G^R$ and $G^{NonR}$ through back-propagating from $L^R$ and $L^{NonR}$, respectively; 
        \STATE $n\defeq n + 1$
        \ENDWHILE
        \IF {$ep > 0$}
        \STATE Update $G_m^R$ and $G_m^{NonR}$ by Eq. \eqref{momentum};
        \ELSE
        \STATE Copy the parameters of $G^R$ and $G^{NonR}$, to $G_m^R$ and $G_m^{NonR}$, respectively;
        \ENDIF
        \STATE $ep\defeq ep + 1$;
        \ENDWHILE
    \end{algorithmic}
\end{algorithm}

\subsection{Robust Collaborative Learning}
As illustrated before, the original target $Y_n$ may contain ambiguous or wrong annotations, which provide mislead optimization direction in training. To avoid the disturbance of noisy labels, we design a collaborative learning algorithm, to explore the robust knowledge across training moments and heterogeneous architectures. Doing so helps to break free from the dependence of the knowledge in pre-trained models, and therefore enables the network to have both high accuracy and fast inference speed. 

Concretely, the knowledge at different training moments throughout the whole training process, can be aggregated by updating the parameters of a momentum network:

\begin{equation}
\begin{aligned}
\textbf{W}_m^{(ep)} \defeq (\textbf{W}_{bp}^{(ep)} + \textbf{W}_m^{(ep-1)}) / 2,
\end{aligned}
\label{momentum}
\end{equation}
where $\textbf{W}_m^{(ep)}$ and $\textbf{W}_{bp}^{(ep)}$ are the parameters of a momentum network, and its corresponding network trained by back-propagation, after the $ep$-th epoch, respectively.

\begin{figure*}[t]
    \centering
    \includegraphics[width=0.16\linewidth]{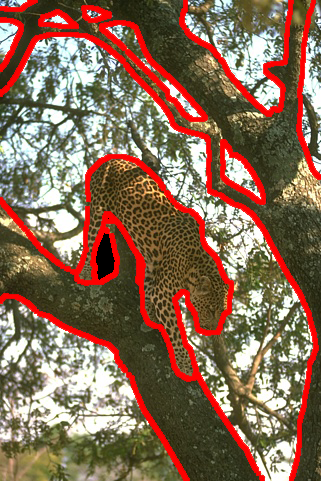}
    \includegraphics[width=0.16\linewidth]{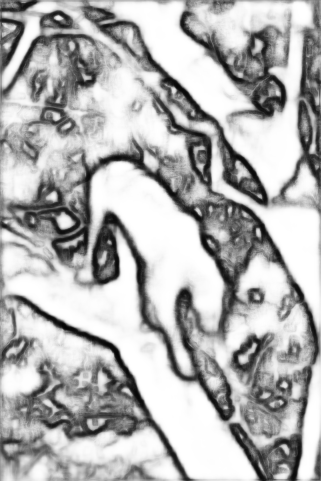}
    \includegraphics[width=0.16\linewidth]{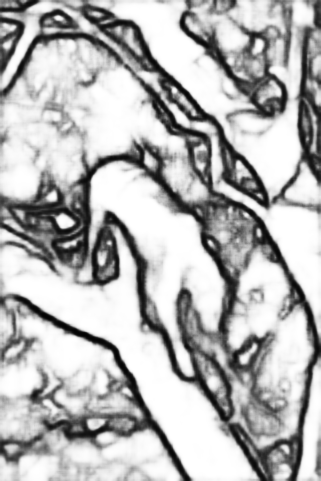}
    \includegraphics[width=0.16\linewidth]{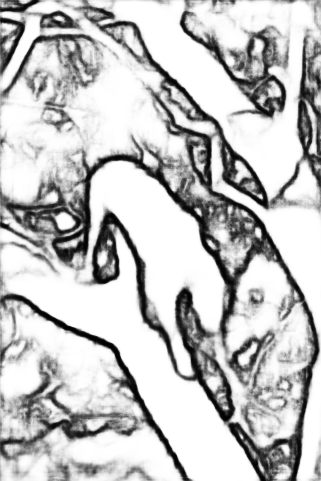}
    \includegraphics[width=0.16\linewidth]{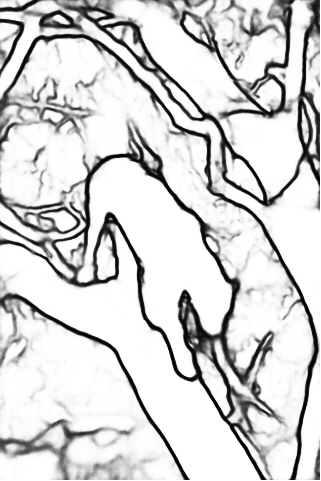}    
    \includegraphics[width=0.16\linewidth]{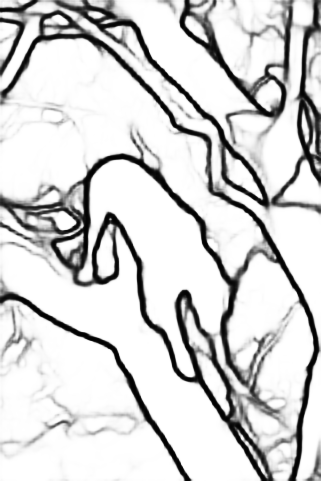}
\\ 

    \includegraphics[width=0.16\linewidth]{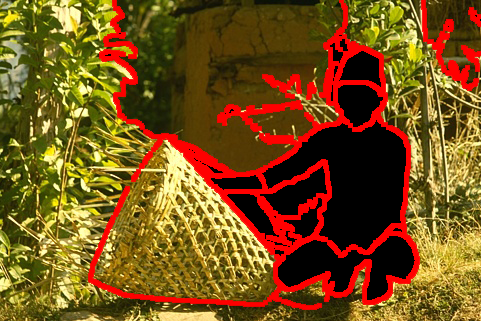}
    \includegraphics[width=0.16\linewidth]{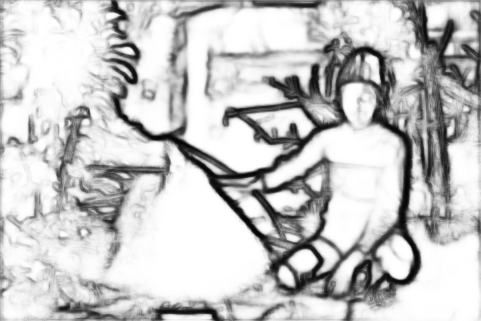}
    \includegraphics[width=0.16\linewidth]{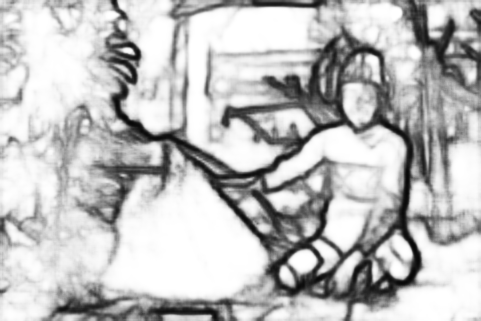}
    \includegraphics[width=0.16\linewidth]{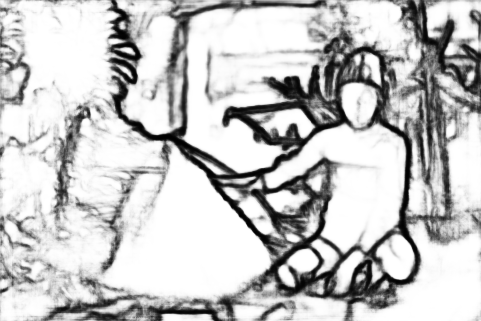}
    \includegraphics[width=0.16\linewidth]{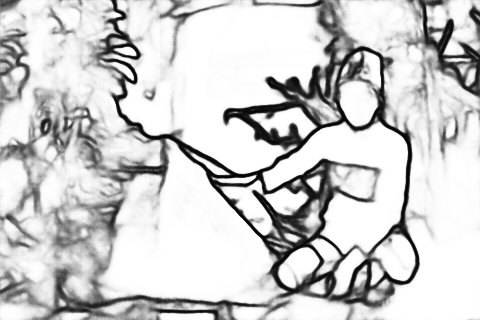}
    \includegraphics[width=0.16\linewidth]{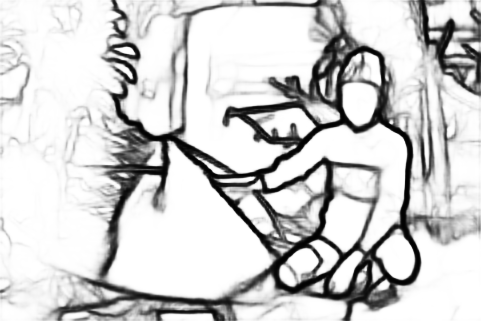}
\\

    \includegraphics[width=0.16\linewidth]{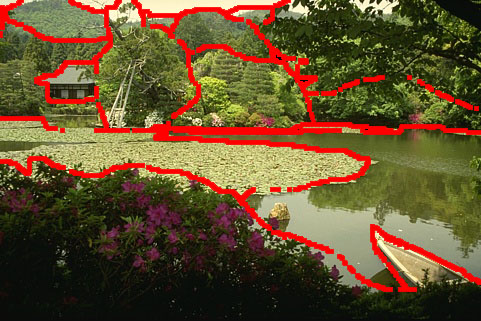}
    \includegraphics[width=0.16\linewidth]{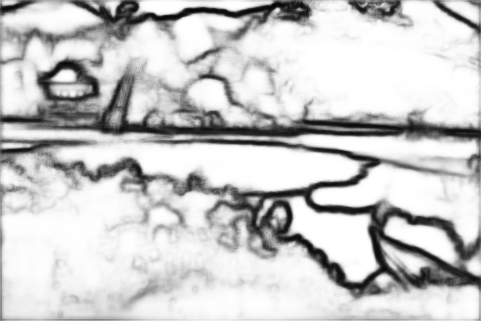}
    \includegraphics[width=0.16\linewidth]{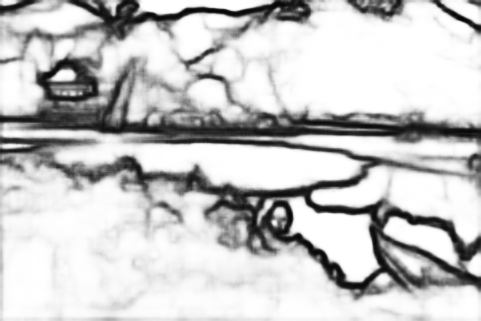}
    \includegraphics[width=0.16\linewidth]{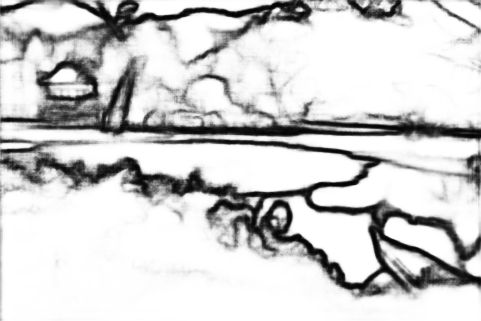}
    \includegraphics[width=0.16\linewidth]{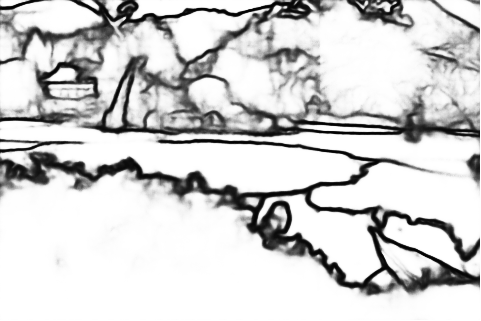}
    \includegraphics[width=0.16\linewidth]{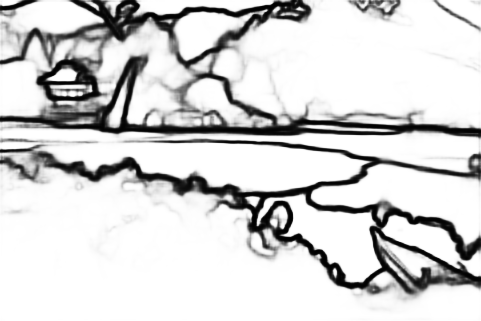}
    
       \quad Input \& GT \quad\quad\quad\quad\quad\quad RCF \quad\quad\quad\quad\quad\quad\quad PiDiNet \quad\quad\quad\quad\quad\quad  BDCN \quad\quad\quad\quad\quad\quad  EDTER \quad\quad\quad\quad\quad\quad PEdger \quad\quad\quad
    
    \caption{Visual comparison with other competitors.}
    \label{com}
\end{figure*}

To combine the knowledge from networks with different structures, we maintain two momentum networks $G_m^R$ and $G_m^{NonR}$ corresponding to the recurrent and non recurrent network $G^R$ and $G^{NonR}$ that trained by back-propagation, respectively. We propose to fuse the predictions of two momentum networks based on the pixel-wise uncertainty of these two models:
\begin{equation}
\begin{aligned}
    M_n \defeq M^{R}_n \circ U^{R}_n + M^{NonR}_n \circ U^{NonR}_n,\\
    \label{fusion}
\end{aligned}
\end{equation}
in which
\begin{equation}
\begin{aligned}
    U^{R}_n\defeq &\frac{|M^{R}_n - 0.5|}{|M^{R}_n - 0.5| + |M^{NonR}_n - 0.5|},\\
    U^{NonR}_n\defeq &\frac{|M^{NonR}_n - 0.5|}{|M^{R}_n - 0.5| + |M^{NonR}_n - 0.5|},
    \label{uncertainty}
\end{aligned}
\end{equation}
where $\circ$ is the Hadamard product, $M^{R}_n\defeq G_m^R(X_n)$ and $M^{NonR}_n\defeq G_m^{NonR}(X_n)$, $|\cdot|$ is pixel-wise absolute operation, $U^{R}_n$ and $U^{NonR}_n$ are the uncertainty scores of the predictions. The closer (farther) the predictions by a network are to 0.5 compared to the other, the more (less) uncertain about the predictions by this network, and thus the less (more) contributions it should make to fusion. It can be seen from Table \ref{ablation} that, the uncertainty-aware weighting method achieves higher edge detection accuracy,  compared to simply averaging the predictions with the same weight. 

Having the robust $M_n$  that integrates the knowledge across training moments and models, the original noisy target $Y_n$ can be replaced with the corrected $\widetilde{Y}_n$, which is calculated through:
\begin{equation}
    \widetilde{Y}_n \defeq \eta_{ep} \cdot M_n  + (1 - \eta_{ep}) \cdot Y_n,
    \label{soft}
\end{equation}
where $\eta_{ep}$ is the weight that controls how much we are going to trust the knowledge from $M_n$, at $ep$-th epoch. It is worth to mention that, $\widetilde{Y}_n$ can also be regarded as the soft target \cite{DBLP:conf/cvpr/DiazM19,DBLP:conf/iccv/ZiZMJ21,teachingwithsoft,DBLP:conf/iclr/ZhouSCZWYZ21,DBLP:conf/ijcai/WangNFHWWLZL22,DBLP:conf/aaai/LienenH21}, which suppresses over-confident class probabilities in the original hard target. In our collaborative setting, since all the models are optimized collaboratively from scratch, they generally do not have enough knowledge about data distributions at the early stage of training. Hence, inspired by the self distillation technique \cite{SelfDistillation}, we gradually increase $\eta_{ep}$ in training, which is computed as:
\begin{equation}
    \eta_{ep} \defeq \eta_{EP} \times \frac{ep}{EP},
    \label{weightincrease}
\end{equation}
where $EP$ is the total number of epoch for training and $\eta_{EP}$ is the $\eta_{ep}$ at the last epoch. In this paper, we set $\eta_{EP} \defeq 0.8$.  The whole process can be summarized  in \textbf{Algorithm} \ref{alg:algorithm}.

\subsection{Loss Function} As for the losses in training, we adopt the widely-used cross-entropy loss on all the side-outputs and final edge maps. The formulation is:
\begin{equation}
\begin{aligned}
l(E_n, \widetilde{Y}_n) \defeq - \alpha \cdot &\sum_{i=1}^{|X_n|} \widetilde{Y}_n^{(i)} \text{ Log}(E_n^{(i)}) - \\
               -  \beta \cdot &\sum_{i=1}^{|X_n|} (1 - \widetilde{Y}_n^{(i)}) \text{ Log}(1 - E_n^{(i)}),
\end{aligned}
\end{equation}
where $\alpha \defeq \lambda \cdot \frac{|Y_n^{+}|}{|Y_n^{+}| + |Y_n^{-}|}$ and $ \beta \defeq \frac{|Y_n^{-}|}{|Y_n^{+}| + |Y_n^{-}|}$. Moreover, $|Y_n^{+}|\defeq||Y_n\circ \widetilde{Y}_n||_1$ and $|Y_n^{-}|\defeq||(1-Y_n)\circ (1-\widetilde{Y}_n)||_1$ denote the positive and negative sets of the corrected ground truth corresponding to the $n$-th sample, respectively. $\lambda$ is used to balance the positive and negative pixels. $|X_n|$ denotes the number of pixels in  the image $X_n$.  $\widetilde{Y}_n^{(i)}$ and $E_n^{(i)}$ refer to the $i$-th pixel in $\widetilde{Y}_n$ and $E_n$, respectively. The final loss functions $L^R$ and $L^{NonR}$ respectively for $G^R$ and $G^{NonR}$, are formulated as:
\begin{equation}
\begin{aligned}
L^R\defeq& l(E_n^R, \widetilde{Y}_n) + \\
\sum_{t=1}^T [l(\sigma(R_n^{f2c(t)}), &\widetilde{Y}_n) + l(\sigma(R_n^{c2f(t)}), \widetilde{Y}_n)],
\end{aligned}
\label{loss1}
\end{equation}
\begin{equation}
\begin{aligned}
L^{NonR}\defeq& l(E_n^{NonR}, \widetilde{Y}_n) + \\
\sum_{t=1}^4 [l(\sigma(O_n^{f2c(t)}), &\widetilde{Y}_n) + l(\sigma(O_n^{c2f(t)}), \widetilde{Y}_n)],
\end{aligned}
\label{loss2}
\end{equation}
where $\sigma(\cdot)$ is the Sigmoid function, $E_n^R\defeq G^R(X_n)$ and $E_N^{NonR}\defeq G^{NonR}(X_n)$. 

\begin{table*}[t]
\centering
\caption{Comparison on the BSDS dataset. The running speeds are tested on an RTX 3090 GPU  with 200 $\times$ 200 images as input if not specified, while $\dagger$ indicates cited GPU speeds. Our results are highlighted in bold.}
\begin{tabular}{c|c|c|cc|cc|c}
\toprule
\multicolumn{2}{c|}{Method} & Pub.'Year & ODS & OIS & Throughput & Params & Pre-training\\
\hline
Human & - & - & .803 & .803 & - & - & - \\
\midrule
Canny \cite{Canny} & & PAMI'86 & .600 & .640 & 31FPS & - & No \\
Pb \cite{pb} &  & PAMI'04 & .672 & .695 &  - & - & No \\
gPb-UCM \cite{gpu-ucm} & Traditional  & PAMI'11 & .726 & .757 &  - & - & No \\
ISCRA \cite{ISCRA} & Methods & CVPR'13 & .717 & .752 & - & - & No\\
SE \cite{se} & & ICCV'13 & .743 & .763 &  12.5FPS$\dagger$ & - & No \\
OEF \cite{oef} & & CVPR'15 & .746 & .770 &  2/3FPS$\dagger$ & - & No \\
\midrule
\midrule
DeepEdge \cite{deepedge} & & CVPR'15  & .753 & .772 & 1/1000FPS$\dagger$ & - & ImageNet \\
DeepContour \cite{deepcontour} & & CVPR'15  & .757 & .776 & 1/30FPS$\dagger$ & 27.5MB & ImageNet \\
HFL\cite{HFL}  & & ICCV'15 & .767 & .788 &  5/6FPS$\dagger$ & - & ImageNet \\
CEDN \cite{CEDN} & & CVPR'16 & .788 & .804 & 10FPS$\dagger$ & - & ImageNet \\
COB \cite{COB} & & PAMI'17 & .793 & .820 &  - & - & ImageNet \\
HED \cite{HED} & CNN & IJCV'17 & .788 & .808 & 73FPS & 14.7MB & ImageNet \\
CED \cite{CED} & based & CVPR'17 & .794 & .811 & - & 14.9MB & ImageNet \\
AMH-Net \cite{AMH} &  Methods & NeurIPS'17 & .798 & .829 & - & - & ImageNet \\
LPCB \cite{LPCB} & & ECCV'18 & .808 & .824 & 30FPS$\dagger$ & - & ImageNet \\
RCF \cite{RCF} &  & PAMI'19 & .806 & .823 & 61FPS & 14.8MB & ImageNet \\
BDCN \cite{BDCN_cvpr} & & CVPR'19 & .820 & .838 & 44FPS & 16.3MB & ImageNet \\
DSCD \cite{DSCD} & & ACMMM'20 & .813 & .836 & - & - & ImageNet \\
LDC \cite{LDC} & & ACMMM'21 & .812 & .826 & - & - & ImageNet \\
FCL \cite{FCL} & & NN'22 & .815 & .834 & - & - & ImageNet \\
\midrule
EDTER \cite{EDTER} & ViT based & CVPR'22 & .832 & .847 & 2.1FPS & 468.84MB & ImageNet \\
\midrule
\midrule
PiDiNet \cite{PiDiNet} & CNN & ICCV'21 & .807 & .823 & 93FPS & 710KB & No \\
\textbf{PEdger} & based  & - & \textbf{.821} & \textbf{.840} & \textbf{92FPS} & \textbf{734KB} & \textbf{No}\\
\textbf{PEdger-large} & Methods & - & \textbf{.823} & \textbf{.841} & \textbf{71FPS} & \textbf{979KB} & \textbf{No}\\
\bottomrule
\end{tabular}
\label{BSDS_result}
\end{table*}

\section{Experimental Validation}
\subsection{Implementation Details}
\label{details}
We implement our network using PyTorch library on RTX 3090 GPU. In detail, PEdger is randomly initialized and trained for 30 epochs using AdamW optimizer. The momentum and weight decay of the optimizer are set to 0.9 and 0.001, respectively. The warm-up strategy is adopted for the training of the first 4 epochs whose learning rates are gradually increased to 0.001, and then decayed in a linear way. The recurrent step $T$ is set to 5 for recurrent models. $\lambda$ is set to 1.1 for BSDS, and 1.3 for NYUD. Given a predicted edge confidence, a threshold is needed to produce the binary edge map. The threshold for binarizing the ground truth edge maps is set to 0.2 for both BSDS500. No threshold is needed for NYUD since the images are singly annotated. The batch size is set to be 16, and is implemented by gradient accumulation since the sizes of images are varied in training. During training, we randomly perturb the brightness, contrast, saturation, and hue of input images, from 50\% to 150\% of their original values, and randomly convert input images to gray-scale images in a probability of 0.2, for data augmentation. When evaluating, standard non-maximum suppression (NMS) \cite{se} is applied to thin detected edges, and both Optimal Dataset Scale (ODS) and Optimal Image Scale (OIS) F-score are reported. The Sigmoid activation is also replaced with $e^{x-0.5}/(e^{x-0.5} + e^{-x+0.5})$ in testing.

\subsection{Datasets \& Competitors}
We compare our method with some non-deep learning algorithms, including Canny \cite{Canny}, Pb \cite{pb}, gPb-UCM \cite{gpu-ucm}, ISCRA \cite{ISCRA}, SE \cite{se}, and OEF \cite{oef}, and some recent deep learning based approaches, including DeepEdge \cite{deepedge}, DeepContour \cite{deepcontour}, HFL \cite{HFL}, CEDN \cite{CEDN}, HED \cite{HED}, COB \cite{COB}, CED \cite{CED}, AMH-Net \cite{AMH}, LPCB \cite{LPCB}, RCF \cite{RCF}, BDCN \cite{BDCN_cvpr}, DSCD \cite{DSCD}, PiDiNet \cite{PiDiNet}, LDC \cite{LDC}, FCL \cite{FCL}, and EDTER (ViT-based) \cite{EDTER}. 

Two datasets employed for evaluation include:
1) The \textit{BSDS500} \cite{gpu-ucm} has 200 training, 100 validation and 200 test images. The label is annotated by 4 to 9 participants. Following previous works, we train our model on the data consisting of augmented BSDS and VOC Context dataset; and 2) The \textit{NYU Depth (NYUD)} dataset \cite{rgbd} contains 1449 RGB and HHA image pairs, which is split into train (381 images), validation (414 images), and test sets (654 images). Following previous works, we increase the maximum tolerance allowed for correct matches of edge predictions to ground truth to .011, when evaluating.

\subsection{Ablation Study}
We conduct our ablation study on the mixture of augmented BSDS and PASCAL VOC \cite{PASCAL_VOC} dataset. 

\paragraph{Effectiveness of integrating across training moments and models.} We report the results of four alternatives: 1) \textbf{Baseline.} Training a model with non recurrent structure supervised by the original noisy targets $Y_n$; 2)  \textbf{No Information from neither different Moments nor heterogeneous Structures (NIMS)}. Generating the targets $\widetilde{Y}_n$ neither based on knowledge from different training moments nor heterogeneous architectures. The knowledge learned by a single model trained after the last epoch, are used to produce $\widetilde{Y}_n$; 3) \textbf{Ensemble only Across Different Training Moments (EADM)}. Generating the targets $\widetilde{Y}_n$ only based on the knowledge across different training moments. In this setting, only the momentum and back-propagation network with non recurrent structure, is maintained in training; 4) \textbf{Ensemble only Across Different Structures (EADS)}. Generating the targets $\widetilde{Y}_n$ only based on the knowledge across different architectures without maintaining momentum networks, where the knowledge learned from the last epoch, are fused to obtain $M_n$. 5) \textbf{Mutual Learning of Heterogeneous Structures (MLHS)}. Similar to \cite{mutual}, the recurrent and non recurrent network will mutually supervise each other. The recurrent model is supervised by the non recurrent model and $Y_n$, while the non recurrent model is supervised by the recurrent one and $Y_n$. In this alternative, the knowledge across training moments is incorporated.

\begin{table}[t]
\centering
\small
\caption{Ablation study on the effectiveness of the proposed collaborative learning strategy on BSDS dataset in terms of ODS-F. All the results are computed with single-scale inputs. }
\begin{tabular}{cccc}
  \hline
Baseline & NIMS & EADS & EADM \\
\hline
.809 & .811 & .814 & .818 \\
  \hline
MLHS & Average & Synchron. & \textbf{Ours} \\
\hline
.818 & .817 & .819 & \textbf{.821}\\

\bottomrule
\end{tabular}
\label{ablation}
\end{table}
\begin{table}[t]
\centering
\caption{Ablation study on the effectiveness of the two network architectures involved in training, on BSDS dataset in terms of ODS-F. All results are computed with single-scale inputs, and our results are highlighted in \textbf{bold}. $\dagger$ denotes the non recurrent structure whose number of parameters decreases as the resolutions of features decrease.}
\begin{tabular}{cc|cc}
  \hline
 \multicolumn{2}{c|}{Network 1} & \multicolumn{2}{c}{Network 2} \\
 \midrule
Architecture & ODS & Architecture  & ODS  \\
\midrule
 Recurrent  & .815 & Recurrent & .816 \\
 Non-Recurrent & .818 & Non-Recurrent & .819 \\
 Recurrent & .816 & Non-Recurrent$\dagger$ & .818 \\
 \textbf{Recurrent} & \textbf{.817} & \textbf{Non-Recurrent} & \textbf{.821} \\
\bottomrule
\end{tabular}
\label{netarch}
\end{table} 
As can be seen from Table \ref{ablation}, the accuracy of Baseline is the worst one among all the alternatives, due to the mislead noises/outliers in manually annotated labels. Though slightly better than the baseline when using $\widetilde{Y}_n$ instead of noisy $Y_n$, NIMS still lags behind other alternatives. In comparison, using the targets $\widetilde{Y}_n$ produced by either EADM or EADS can evidently improve the performance, proving the necessity of either training moments or heterogeneous architectures, in solving the problem of noisy annotations. When combining the knowledge from both of these two aspects (\emph{Ours}), the best result is achieved, benefiting from the robust information that resists the disturbance of noises. It is worth mentioning that, MLHM is inferior to our method, which shows the effectiveness of fusing the predictions of recurrent and non recurrent networks using Eq. \ref{fusion}.

\begin{table}[t]
\centering
\caption{Comparison on the NYUD dataset. All results are computed with single-scale RGB inputs. }
\begin{tabular}{c|c|cc|c}
  \hline
 Methods & Pub.'Year & ODS & OIS & Throughput \\
\midrule
 gPb-UCM \cite{gpu-ucm} & PAMI'10 & .632 & .661 & 1/360FPS$\dagger$\\
 SE \cite{se} & ICCV'13 & .695 & .708 & 5FPS$\dagger$ \\
  OEF \cite{oef} & CVPR'15 & .651 & .667 & -\\
\midrule
HED \cite{HED} & IJCV'17 & .720 & .734 & 67FPS\\
LPCB \cite{LPCB} & ECCV'18 & .739 & .754 & - \\
RCF \cite{RCF} & PAMI'19 & .743 & .757 & 58FPS \\
AMH-Net \cite{AMH} & NeurIPS'17 & .744 & .758 & - \\
BDCN \cite{BDCN_cvpr} & CVPR'19 & .748 & .763 & 35FPS \\
EDTER \cite{EDTER} & CVPR'22 & .774 & .789 & 2.2FPS \\
\midrule
PiDiNet \cite{PiDiNet} & ICCV'21 & .733 & .747 & 62FPS$\dagger$ \\
\textbf{PEdger} & - & \textbf{.742} & \textbf{.757} & \textbf{92FPS} \\
\bottomrule
\end{tabular}
\label{NYUD}
\end{table}

\paragraph{Two-Training stage.} To compare the performance of two-stage training, we maintain a momentum network with recurrent structure and its corresponding back-propagation network in the first training stage, where $\widetilde{Y}_n$ is produced solely based on the predictions of the momentum recurrent network. In the second stage, the momentum and back-propagation network with non recurrent structure is updated, while the networks with recurrent structure are frozen. The target $\widetilde{Y}_n$ for supervising the training of second stage, is produced according to Eq \ref{fusion} and Eq. \ref{soft}. The only difference between the second training stage and our original collaborative learning, is whether the networks with recurrent structure are updated. The quantitative result of two-stage learning (\emph{Synchron.} in Table \ref{ablation}) is 0.819, being slightly worse than ours. The reason may be that the performance of recurrent  networks is fixed during the second stage, which limits the improvement of non recurrent networks.

\paragraph{Effectiveness of network architecture design.}  We conduct experiments to see how the choice of two network architectures in training affect performance. As can be seen from the first two rows in Table \ref{netarch}, involving the same network architectures in training, \textit{e.g.}, all networks are with recurrent or non recurrent structure, leads to the sub-optimal performance. It shows that using two discrepant network structures (recurrent \& non recurrent in this work) is critical to capture the diversified knowledge for improving the robustness and accuracy.

In addition, as we have mentioned in Section \ref{NetworkArchitecture}, the amount of parameters for the non recurrent network, should gradually increases as the features size decreases.  To verify the effectiveness of this design, we allocate more (less) parameters for earlier (later) stages with larger (smaller) feature sizes for comparison. Besides slower than our original design, it can be seen that the accuracy of both recurrent and non recurrent network is inferior to ours  (recurrent: 0.817 vs. 0.816, non recurrent: 0.821 vs. 0.818)  when the parameter amount decreases for later stages as denoted by $\dagger$. The reason lies in that later stages for learning semantic higher-level information, require more parameters than eariler stages for lower-level details. 

\paragraph{Effectiveness of uncertainty-aware fusion.} In this section, we conduct the experiment to validate the effectiveness of our proposed uncertainty-aware combination strategy, as formulated in Eq. \ref{fusion}. To compare between with and without considering the uncertainties, we alternatively average the predictions of recurrent and non recurrent networks using the same weight, that is to say, $M_n$ is produced through $(M^R_n+M^{NonR}_n) / 2$. The quantitative result of simply averaging the predictions is 0.817 (\emph{Average} in Table 2), which is lower than our method in a large margin. It verifies that the larger the confidence of the model in its prediction, the more contribution it should make when emsembling. Otherwise, the low-confident predictions, which are highly likely to be wrong, will cause harmful impacts in fusion.

\begin{figure}[t!]
    \centering
    \includegraphics[width=\linewidth]{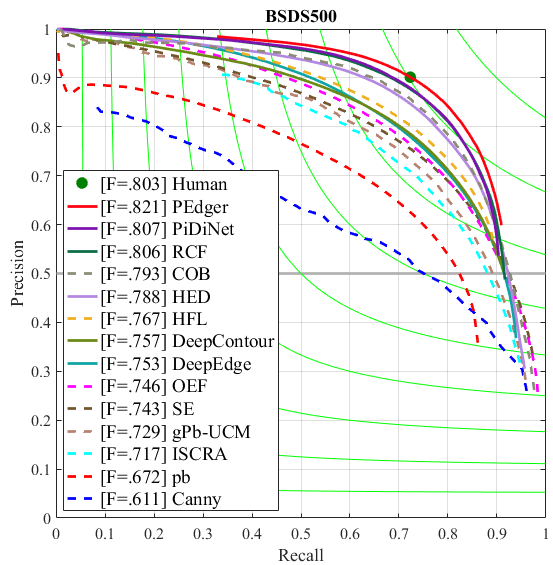}
    \caption{Precision-Recall curves of our method and other competitors on the BSDS dataset.}
    \label{pr}
\end{figure}

\subsection{Comparison with State-of-the-arts}
It can be seen from Fig. \ref{pr} and Tab. \ref{BSDS_result} that our PEdger can achieve high accuracy on BSDS, {\it i.e.}, with ODS-F of 0.821 and OIS-F of 0.840. Though inferior to EDTER, our PEdger runs about 49$\times$ faster than it with merely 734KB parameters. Notice that the two-stage training is required for the global and local transformers of EDTER, making its training procedure inefficient. For the CNN based approaches, our PEdger outperforms the recently proposed BDCN while still achieving nearly 100 FPS. Compared with the only work for efficient edge detection, {\it i.e.}, PiDiNet, we have a significant advantage in terms of ODS/OIS (ODS: 0.821 vs. 0.807, OIS: 0.840 vs. 0.815). 
We also show the visual results in Fig. \ref{com}. It can be seen that our results are much more clearer than those of other competitors, with less superfluous aesthetic predictions in the non edge regions.

 The results on NYUD are shown in Tab. \ref{NYUD}. We can see that our method is able to achieve highly comparable results among the state-of-the-art methods while being efficient. Notice that we adopt the identical network architectures and hyper-parameters for all  datasets if not otherwise specified.

\section{Conclusion}
We proposed a novel collaborative learning framework, namely PEdger, to  simultaneously address the 1) dependence on a large amount of pre-trained parameters, and 2) the performance degradation caused by noisy labels. It is able to integrate the knowledge from different training moments in the whole training procedure, and heterogeneous network architectures. Moreover, we customized an efficient and compact network, contributing to fast and accurate edge detection. Our PEdger can be trained from scratch and surpass human-level performance, breaking the dependence of backbone pre-trained on large scale datasets. Experimental results have revealed the advances of our proposed PEdger over the state-of-the-arts, making it attractive for  real-time applications. 

\section*{Acknowledgements}
This work was supported by National Natural Science Foundation of China under Grant 62072327.

\bibliographystyle{ACM-Reference-Format}
\balance
\bibliography{reference}


\end{document}